\documentclass{article}
\usepackage{spconf,amsmath,graphicx}

\usepackage{float}
\usepackage[linesnumbered,boxruled]{algorithm2e}
\usepackage{verbatim}
\usepackage{tikz,amsmath}

\makeatletter
\newcommand{\removelatexerror}{\let\@latex@error\@gobble}
\makeatother

\newsavebox\IBoxA \newsavebox\IBoxB \newlength\IHeight
\newcommand\TwoFig[6]{
  \sbox\IBoxA{\includegraphics[width=0.45\textwidth]{#1}}
  \sbox\IBoxB{\includegraphics[width=0.45\textwidth]{#4}}%
  \ifdim\ht\IBoxA>\ht\IBoxB
    \setlength\IHeight{\ht\IBoxB}%
  \else\setlength\IHeight{\ht\IBoxA}\fi
  \begin{figure}[!htb]
  \minipage[t]{0.45\textwidth}\centering
  \includegraphics[height=\IHeight]{#1}
  \caption{#2}\label{#3}
  \endminipage\hfill
  \minipage[t]{0.45\textwidth}\centering
  \includegraphics[height=\IHeight]{#4}
  \caption{#5}\label{#6}
  \endminipage 
  \end{figure}%
}

\makeatletter
\newcommand{\algorithmfootnote}[2][\footnotesize]{%
\removelatexerror
  \let\old@algocf@finish\@algocf@finish
  \def\@algocf@finish{\old@algocf@finish
    \leavevmode\rlap{\begin{minipage}{\linewidth}
    #1#2
    \end{minipage}}%
  }%
}
\makeatother

\title{Enhancing the Regularization Effect of Weight Pruning in Artificial Neural Networks}
\name{Brian Bartoldson$^{\star}$ \qquad Adrian Barbu$^{\dagger}$ \qquad Gordon Erlebacher$^{\star}$}

\address{Florida State University \\ $^{\star}$ Department of Scientific Computing, $^{\dagger}$Department of Statistics \\ E-mail for correspondence: bbartoldson@fsu.edu}

\begin{document}
%

\newcommand\hlight[1]{\tikz[overlay, remember picture,baseline=-\the\dimexpr\fontdimen22\textfont2\relax]\node[rectangle,fill=blue!50,rounded corners,fill opacity = 0.3,text opacity =1] {$#1$};} 

\maketitle
\begin{abstract}
Artificial neural networks (ANNs) may not be worth their computational/memory costs when used in mobile phones or embedded devices. Parameter-pruning algorithms combat these costs, with some algorithms capable of removing over 90\% of an ANN's weights without harming the ANN's performance. Removing weights from an ANN is a form of regularization, but existing pruning algorithms do not significantly improve generalization error. We show that pruning ANNs can improve generalization if pruning targets large weights instead of small weights. Applying our pruning algorithm to an ANN leads to a higher image classification accuracy on CIFAR-10 data than applying the popular regularizer dropout. The pruning couples this higher accuracy with an 85\% reduction of the ANN's parameter count.
\end{abstract}
\section{Introduction}
\label{sec:intro}

Artificial neural networks (ANNs) attain state-of-the-art performance on a variety of computer vision problems. The capacity of an ANN to solve a problem is a function of the ANN's size, which can be roughly measured by its number of connection-weight parameters \cite{lai2018not}. The fact that ANNs commonly use more than $10^6$ 32-bit weights exacerbates the following deployment issues (especially on mobile phones and embedded devices): 1) the memory required to store and run the ANN is limited; and 2) the ANN introduces a computational burden that can be costly in terms of time and power consumption \cite{narang2017exploring, han2015deep}. 

Among other approaches, pruning a large fraction of connection weights has helped mitigate ANN size burdens \cite{narang2017exploring, han2015deep, yu2012exploiting, han2015learning}. In 2012, Yu et al.  \cite{yu2012exploiting} observed that roughly 70\% of weights in a trained ANN had magnitudes less than 0.1, and that they could prune these weights without significantly altering the ANN's accuracy. More recent work \cite{narang2017exploring} demonstrated that weight pruning could reduce an ANN's parameter count by 90\% and speed-up an ANN's inference/forward-pass calculations sevenfold. Parameter pruning is particularly useful because it can be combined with other compression strategies like ANN weight quantization \cite{han2015deep, wu2016quantized}.

Pruning parameters from a machine learning model is also done to reduce overfitting to the training data. Barbu et al.  \cite{barbu2017feature} found that pruning model parameters according to an annealing schedule led to better generalization on computer vision tasks than using $l^1$ or $l^2$ regularization. These results inspired our annealed pruning (AP) algorithm, which can shrink ANN parameter counts by roughly 90\% and simultaneously improve test-set accuracies to a greater degree than dropout. 

\subsection{Related Work}
\label{sec:related}

ANN regularization algorithms like dropout \cite{hinton2012improving} restrict a neural network's ability to overfit to its training data, thereby empowering higher test-set accuracies (a proxy for generalization). Since dropout was introduced in 2012, it has facilitated state-of-the-art results on multiple computer vision benchmarks \cite{hinton2012improving}. Dropconnect, a generalization of dropout, works by temporarily setting to zero a random subset of weights on each forward pass \cite{wan2013regularization}. Our algorithm, annealed pruning (AP), is similar to dropconnect in that both methods create random subsets of weights for different forward passes. Unlike dropconnect, AP ultimately removes a large subset of weights to permanently reduce an ANN's size.

Pruning ANN weights to achieve regularization has been considered since at least the Optimal Brain Damage work in 1990 \cite{lecun1990optimal}. Despite that fact, modern pruning algorithms \cite{narang2017exploring, han2015deep, yu2012exploiting, han2015learning} do not significantly improve the pruned ANN's generalization. An example of a basic ANN weight-pruning approach is shown in algorithm \ref{sparseDNN}.

\newcommand{\pushline}{\Indp}
\newcommand{\popline}{\Indm\dosemic}
\let\oldnl\nl
\newcommand{\nonl}{\renewcommand{\nl}{\let\nl\oldnl}}

\begin{algorithm}
\label{sparseDNN}
\caption{Main Steps to Train a Sparse ANN \cite{yu2012exploiting}.}
\SetAlgoLined
Train a fully connected ANN.

Keep only the connections whose weight magnitudes are in top
$q$.

Continue training the ANN with the sparseness pattern generated from Step 2 unchanged.

\algorithmfootnote{$q$ is the maximal number of non-zero parameters allowed in the final ANN. }
\end{algorithm}

Current pruning approaches target the weights with the smallest magnitudes, or the weights with the least importance to maintaining the unpruned ANN's loss (variable importance can be measured via a Taylor series of the loss function) \cite{narang2017exploring, han2015deep, yu2012exploiting, han2015learning,lecun1990optimal}. AP grants generalization improvements when pruning small weights, but AP gives significantly better test-set accuracy gains when pruning the weights with the largest magnitudes. We suspect that this phenomenon is driven by a regularizing mechanism present in dropout/dropconnect. Namely, dropout pressures learned features to be generally useful by constantly changing the subset of features used by the ANN \cite{hinton2012improving}. Intuitively, pruning connection-weights (e.g. the weights connecting layers  $l_{i-1}$ and $l_i$) during training makes less useful those features in layer $l_{i-1}$ that rely on the presence of particular connections to neurons in layer $l_i$, pressuring features learned in layer $l_{i-1}$ to generalize well. We speculate that pruning larger weights strengthens this force on features to be robust.

The basis of the AP algorithm introduced here is the feature-selecting algorithm \ref{FSA} from Barbu et al. \cite{barbu2017feature}, which is designed to improve a machine learning model's generalization by pruning parameters. We make several modifications to algorithm \ref{FSA} to construct AP. As mentioned above, we prune large weights rather than small weights. Also, AP allows pruned weights a chance to reenter the ANN. Narang et al. \cite{narang2017exploring} allows a pruned weight back into the network if (roughly speaking) the derivative of the loss with respect to the pruned weight is larger than the threshold for pruning. AP allows weights back into the network in a randomized fashion (described in section \ref{sec:AP}) that was inspired by the application of genetic algorithms to neural networks in \cite{stanley2002evolving}.

\begin{algorithm}
\label{FSA}
\SetAlgoLined
\SetKwInOut{Input}{Input}
\SetKwInOut{Output}{Output}
\Input{Training examples $\{(x_i,y_i)\}_{i=1}^N$}
\Output{Trained classifier parameter vector $w$}
Initialize $w$

 \For{e = 1 to $N^{iter}$}{
  Update $w \leftarrow w - \eta \frac{\partial L(w)}{\partial w}$
  
  $M_e = k + (M - k)\max( 0, \frac{N^{epochs} - 2e}{2e \mu + N^{iter}})$
  
  Keep the $M_e$ weights with the highest 
  
  \pushline\nonl absolute value
 }
 \caption{Feature Selection with Annealing (FSA) \cite{barbu2017feature}.}
 
 \algorithmfootnote{ $M=|w|$, $\eta$ is the learning rate parameter, and $\mu$ is a parameter that alters the rate at which connections are removed from the network . }
\end{algorithm}

Our experiments use AP to prune parameters from convolutional neural networks (CNNs), a type of ANN that is widely used for computer vision tasks. AP targets a CNN's dense layers, which comprise the  majority of parameters in many CNN architectures. For example, over 95\% of AlexNet's parameters are in its dense layers \cite{krizhevsky2012imagenet}. Counterexamples include ResNet architectures \cite{he2016deep}, which can have fewer than 5\% of their parameters in fully-connected layers \cite{wen2016learning}. Architectures like ResNet could motivate a use for AP in conjunction with a convolutional-filter-pruning approach such as those taken in  \cite{wen2016learning} and \cite{molchanov2016pruning}.

\section{Annealed Pruning}
\label{sec:AP}

Here we explain the annealed pruning (AP) algorithm visually with matrices, and then with the pseudocode in algorithm \ref{AP}.

AP's main inputs are the ANN layer targeted for pruning, and the desired percentage of weights to keep in that layer's weight matrix $W$. At the end of each ANN training epoch, we run an AP iteration. On the first iteration, we construct a binary mask matrix $M$ of the same dimensions as $W$, illustrated in equation \ref{AP_eq1}: 

\begin{align}
\label{AP_eq1}
\begin{smallmatrix}\textrm{Weight Matrix, $W$}\\
\begin{bmatrix}
w_{11} & w_{12} & \cdots & w_{1n} \\
w_{21} & w_{22} & \cdots & w_{2n} \\
\vdots & \vdots & \ddots & \vdots \\
w_{n1} & w_{n2} & \cdots & w_{nn} 
\end{bmatrix}
\end{smallmatrix}
\begin{smallmatrix}\textrm{Mask Matrix, $M$}\\
\begin{bmatrix}
0 & 0 & \cdots & 0 \\
0 & 0 & \cdots & 0 \\
\vdots & \vdots & \ddots & \vdots \\
0 & 0 & \cdots & 0 \\
\end{bmatrix}
\end{smallmatrix}
\end{align}

\noindent
We select a subset of $W$ for pruning. The subset is $\{w_{ij} : \mathrm{abs}(w_{ij}) > t\}$, where $t$ is the weight-magnitude threshold that ensures that the scheduled number of non-zero parameters on that iteration will be met.

\begin{align}
\label{AP_eq2}
\begin{smallmatrix}\textrm{$W$ with Selections}\\
\begin{bmatrix}
w_{11} & \hlight{w_{12}} & \cdots & w_{1n} \\
w_{21} & w_{22} & \cdots & \hlight{w_{2n}} \\
\vdots & \vdots & \ddots & \vdots \\
\hlight{w_{n1}} & w_{n2} & \cdots & w_{nn} 
\end{bmatrix}
\end{smallmatrix}
\end{align}

\noindent
AP adds ones to $M$ in the positions corresponding to the selected weights. Note that some elements of $M$ may already be set to one from the previous AP iteration.

\begin{align}
\label{AP_eq3}
\begin{smallmatrix}\textrm{$M_n$}\\
\begin{bmatrix}
0 & 0 & \cdots & 0 \\
1 & 0 & \cdots & 0 \\
\vdots & \vdots & \ddots & \vdots \\
0 & 0 & \cdots & 0 \\
\end{bmatrix}
\Rightarrow
\end{smallmatrix}
\begin{smallmatrix}\textrm{$M_{n+1}^\prime$}\\
\begin{bmatrix}
0 & \hlight{1} & \cdots & 0 \\
1 & 0 & \cdots & \hlight{1} \\
\vdots & \vdots & \ddots & \vdots \\
\hlight{1} & 0 & \cdots & 0 \\
\end{bmatrix}
\end{smallmatrix}
\end{align}

\noindent
$\{w_{ij} : m_{ij} = 1\}$ are set to zero. That is, $W_{n+1} = W_n \odot \neg M_{n+1}^\prime$. The zeroed $w_{ij}$ remain trainable in the ANN. A subset of $\{m_{ij} : m_{ij} = 1\}$ is randomly chosen and set to 0 to allow $w_{ij}$ multiple chances to be in the final ANN:

\begin{align}
\label{AP_eq4}
\begin{smallmatrix}\textrm{$M_{n+1}^\prime$}\\
\begin{bmatrix}
0 & 1 & \cdots & 0 \\
1 & 0 & \cdots & 1 \\
\vdots & \vdots & \ddots & \vdots \\
1 & 0 & \cdots & 0 \\
\end{bmatrix}
\Rightarrow
\end{smallmatrix}
\begin{smallmatrix}\textrm{$M_{n+1}$}\\
\begin{bmatrix}
0 & 1 & \cdots & 0 \\
1 & 0 & \cdots & 1 \\
\vdots & \vdots & \ddots & \vdots \\
\hlight{0} & 0 & \cdots & 0 \\
\end{bmatrix}
\end{smallmatrix}
\end{align}

\noindent
After equation \ref{AP_eq4}, we resume training the ANN. The ANN trains for $N$ epochs, and pruning takes place on $n<N$ epochs. On the final pruning epoch, $M$ is used to prune $W$ such that $\{w_{ij} : m_{ij} = 1\}$ are no longer trainable and permanently zero.

AP can be implemented as a callback function that runs at the end of an ANN training epoch. Algorithm \ref{AP} gives the pseudocode for AP: 

\begin{algorithm}
\SetKwInOut{Input}{Input} 
\label{AP}
\SetAlgoLined
\caption{Annealed Pruning (AP)}
\Input{$layer$ = ANN layer to prune, $p$ = fraction of weights to keep (we use 0.1), $\mu =$ pruning rate (we use 1), $start =$ first pruning epoch (we use 3), $post =$ post-pruning epochs (we use 3), $e$ = current epoch, $N$ = \# of epochs to train ANN}

$\mathbf{w} = layer$'s weights

\If{$e==1$}{

  $M = |\mathbf{w}|$ \tcp{\# of weights in $layer$}
  
  $k = p*M $\tcp{\# of weights to keep}

  $\mathbf{mask} = 0$ \tcp{the mask has length $M$}
  
  $n = N-post - start + 1$ \tcp{\# AP epochs}
}

$i = e-start + 1$ \tcp{AP iteration tracker}

\If{$n \geq i \geq 1$}{

	$M\_e  =  k + (M - k) \max( 0, \frac{n - i}{2i \mu + n})$ 
    
    $pruned = \{\mathbf{w}_j\ : \mathbf{mask}_j = 1 \}$ 

	$pruned\_ct = |pruned|$
    
    $unpruned = \{\mathbf{w}_j\ : \mathbf{mask}_j = 0 \}$

	$unpruned\_ct = |unpruned|$
    
    $threshold_p = \frac{M_e}{unpruned\_ct}$
    
    Update $threshold \leftarrow$ the $threshold_p$ percentile of abs$(unpruned)$
    
    Update $\mathbf{mask}_j \leftarrow 1$ where abs$(\mathbf{w}_j)>threshold$

	Update $layer$'s weights $\leftarrow \mathbf{w}  \odot \neg \mathbf{mask}$    
    
    Update $\mathbf{mask}_q \leftarrow 0$ where $q \subset \{j:\mathbf{mask}_j=1\}$
 }

\If{$i == n$}{
	Permanently prune $layer$'s weights such that $\{\mathbf{w}_j : \mathbf{mask}_j =1\}=0$
}
  \algorithmfootnote{On line 10, we calculate the scheduled number of nonzero weights at the end of epoch $e$. On line 19, we unmask a random sample of size $pruned\_ct * a$, where $a \sim \text{U}(0,b)$, and $b<1$ goes to $0$ as $i \rightarrow n$.}
\end{algorithm}

Line 19 of algorithm \ref{AP} unselects a fraction of the weights previously selected for pruning. The weights that are unselected and the magnitude of the fraction are random, but the number of weights unselected goes to zero as the final pruning epoch is approached. The unselection allows important connections to have multiple chances to be in the network. Unselected connections are still susceptible to pruning at a later iteration.

\section{Experiments}
\label{sec:exp}

CIFAR-10 is a collection of 60,000 $32\times32\times3$ pixel images with 10 classes \cite{cifar_data}. 50,000 images are labeled as model training data, and 10,000 images are labeled as model testing data. Our experiments utilize the training data to train three types of ANN: a baseline convolutional neural network (CNN), the baseline CNN trained with dropout on the inputs to the first dense layer, and the baseline CNN trained with annealed pruning (AP) on the weights of the first dense layer. 

The baseline CNN architecture is described in table \ref{cifar_model}. All convolutional and dense layers are followed by a ReLU activation function except for the last dense layer, which is followed by a softmax classifier. The convolutional layers have 32, 32, 64, and 64 filters, respectively. All filters are $3\times 3$. The max pooling layers pool $2\times 2$ windows. For the dropout CNN, we use a dropout fraction of 0.25 on the set of features that feed into the first dense layer.

\begin{table}[!htbp]
\centering
\begin{tabular}{ l r r }
    \hline
    Layer & Parameters & Parameters after Pruning\\ \hline      
conv2D             &      896      &      896     \\ 
conv2D            &       9,248     &       9,248        \\
max\_pooling2D  &        0       &        0      \\
conv2D          &        18,496    &        18,496       \\
conv2D          &         36,928    &         36,928      \\
max\_pooling2D &       0       &       0      \\
dense                  &           1,180,160   &       $\approx$118,000     \\
dense             &              5,130    &              5,130       \\
\textbf{total}      &    \textbf{1,250,858}     &   \textbf{188,714}        
\end{tabular}
\caption{The baseline CIFAR-10 CNN, its parameter counts by layer, and the parameter counts after pruning with AP.}
\label{cifar_model}
\end{table}

\subsection{AP Improves Generalization More than Dropout}

\begin{figure}[H]
\centering
\includegraphics[width=1\linewidth]{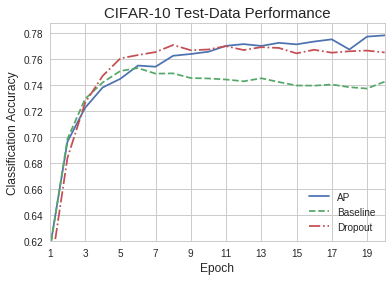}
\caption{CIFAR-10 test-data accuracies for the baseline CNN described in table \ref{cifar_model}, the baseline+dropout, and the baseline+AP. We plot averages of 10 runs for each CNN.}
\label{cifar_test}
\end{figure}

\begin{figure}[H]
\centering
\includegraphics[width=1\linewidth]{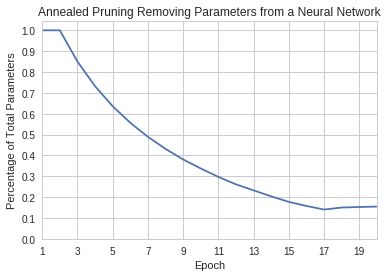}
\caption{The fraction of parameters that were nonzero at epoch end as the CNN trained with AP. The parameters targeted by AP for pruning were not permanently set to zero until epoch 17.}
\label{cifar_prune}
\end{figure}

To ascertain the ability of AP to assist with generalization while removing parameters, we use the CIFAR-10 testing data to compare the image classification accuracies of the three trained CNNs. Figure \ref{cifar_test} shows that the best generalization is achieved with AP, and figure \ref{cifar_prune} shows the pruning schedule that AP uses to remove  85\% of the CNN's weights. 

The baseline CNN starts to overfit (losing test accuracy in favor of training accuracy) around epoch 5. The AP CNN's test data classification accuracy steadily improves throughout the 20 epoch training period. One exception is epoch 17, when the parameters targeted by AP's mask are permanently set to zero--we believe this temporary dip is caused by the CNN struggling with the abrupt loss of model capacity. 

AP improves its baseline model's accuracy by more than preexisting pruning approaches have improved their baselines. The test-data accuracy with AP (78\%) is more than 5\% higher than the baseline CNN accuracy (74.2\%), and higher than the dropout CNN accuracy (76.5\%).


\subsection{Pruning Larger Weights Improves Generalization}
\begin{figure}[H]
\centering
\includegraphics[width=1\linewidth]{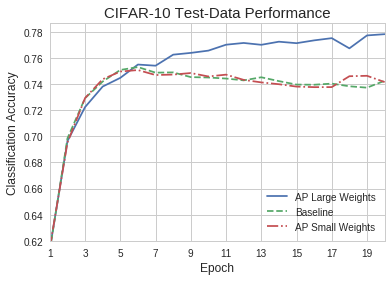}
\caption{CIFAR-10 test-data accuracies are higher when AP targets the largest weights for pruning as opposed to the smallest weights. }
\label{cifar_small_large}
\end{figure}

We modified algorithm \ref{AP} to make AP remove the smallest (or least important) weights in the CNN, which is the approach taken in the pruning literature \cite{narang2017exploring, han2015deep, yu2012exploiting, han2015learning,lecun1990optimal}. AP targeting small weights fails to significantly enhance the generalization of the baseline CNN (figure \ref{cifar_small_large}). This suggests that pruning the largest weights in an ANN leads to better generalization than pruning the smallest weights.

\section{Discussion and Future Work}
\label{sec:discussion}

Like the pruning algorithm in \cite{narang2017exploring}, AP has several hyperparameters. We did not test a wide range of hyperparameters, so our results could improve with different hyperparameter settings. In future work, we will try to understand the sensitivity of our results to these hyperparameters, and develop a heuristic for choosing the appropriate values.

Additionally, we would like to know whether AP works well alongside other regularizers. For instance, could AlexNet, which uses dropout, obtain higher accuracies when trained with AP? 

Lastly, AP allows weights that were selected for pruning to contribute to the network until the last pruning iteration. We are interested in modifying AP to permanently remove weights as soon as they are selected. If this allowed sparse matrices to be used during ANN training, then AP could speed up ANN training in addition to its other benefits.

\section{Conclusion}
\label{sec:conclusion}

Annealed pruning (AP) can reduce ANN parameter counts by almost 90\% and improve generalization more than dropout. In an image-classification experiment with CIFAR-10 data, AP pruned 85\% of a CNN's parameters while generating a 5.1\% improvement in the CNN's test-data accuracy. Our work contributes to the  pruning literature in several ways: 1) we introduce the notion of targeting important ANN connections (large weights) for pruning; 2) we provide evidence that pruning more important connections leads to better generalization; and 3) our algorithm, which combines an annealing schedule with randomized weight reentry, facilitates competitive compression and unsurpassed pruning-based generalization improvements.


\bibliographystyle{IEEEbib}
\bibliography{biblio}

\end{document}